\documentclass[twocolumn,10pt]{asme2e}

\usepackage{psfrag}
\usepackage{graphicx}
\usepackage{epstopdf}
\usepackage{amsmath,amssymb, amsfonts}

\confshortname{IDETC/CIE 2014}
\conffullname{the ASME 2014 International Design Engineering Technical Conferences \&\\
 Computers and Information in Engineering Conference}
\confdate{August 17-20}
\confyear{2014}
\confcity{Buffalo}
\confcountry{USA}
\papernum{DETC2014/MR-34593}

\title{Multi-Objective Design Optimization of the Leg Mechanism for a Piping Inspection Robot}

\author{Renaud HENRY, Damien CHABLAT, \\  {\tensfb Mathieu POREZ, Fr\'ed\'eric BOYER}
 \affiliation{
	Institut de Recherche en Communications et \\
	Cybern\'etique de Nantes (IRCCyN)\\
	UMR CNRS n$^\circ$ 6597\\
	1, rue de la No\"e, 44321 Nantes Cedex 03, France\\
 Email addresses: \\
		\{renaud.henry, damien.chablat\}@irccyn.ec-nantes.fr \\
		\{mathieu.porez, frederic.boyer\}@irccyn.ec-nantes.fr} 
}
\author{Daniel KANAAN
			\affiliation{
	 AREVA NC, Bagnols sur C\`eze, France\\
 Email addresses: daniel.kanaan@areva.com}
}

\begin{document}
\maketitle
\thispagestyle{empty}
\pagestyle{empty}

\begin{abstract}
{\it This paper addresses the dimensional synthesis of an adaptive mechanism of contact points ie a leg mechanism of a piping inspection robot operating in an irradiated area as a nuclear power plant. This studied mechanism is the leading part of the robot sub-system responsible of the locomotion. Firstly, three architectures are chosen from the literature and their properties are described.
 Then, a method using a multi-objective optimization is proposed to determine the best architecture and the optimal geometric parameters of a leg taking into account  environmental and design constraints.
 In this context, the objective functions are the minimization of the mechanism size and the maximization of the transmission force factor. Representations of the Pareto front versus the objective functions and the design parameters are given. Finally, the CAD model of several solutions located on the Pareto front are presented and discussed.}
\end{abstract}

\section*{INTRODUCTION}
\begin{figure}[htb!]
	\centering
		\includegraphics[width=7cm]{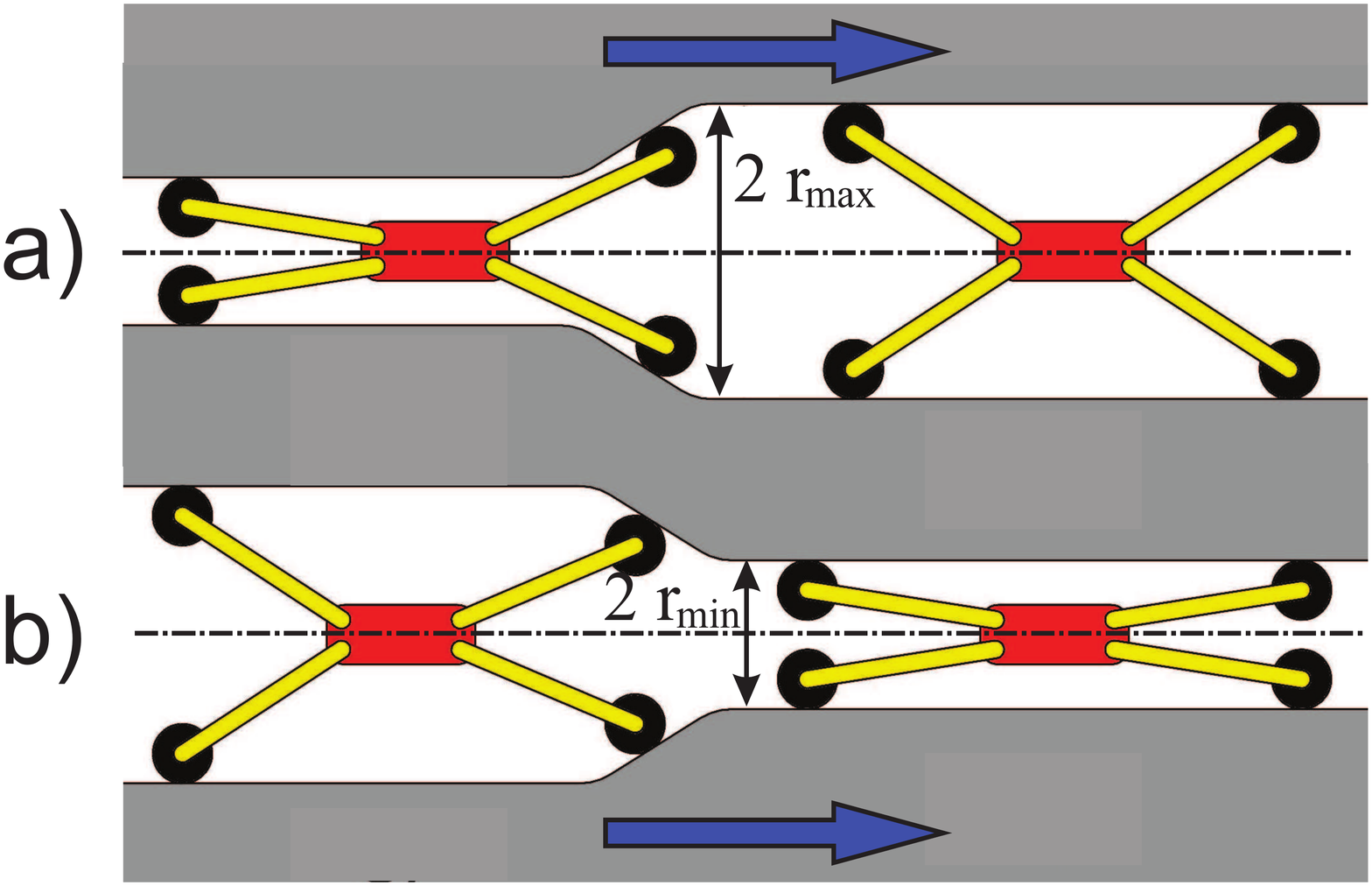}
	\caption{Piping robot with a generic adaptive mechanism of the contact points when the inner diameter: a) increases or b) decreases.}
	\label{fig:robot_pipe}
\end{figure}

In a nuclear power plant, there are many places that the human workers cannot reach due to the high level of irradiation (which can be deadly). However, for safety reasons inherent to a nuclear power plant, the pipe-line equipments (which are large in such a plant) require periodic and rigorous inspections. In this context, the development of robotic system suitable to finding and to repaire a failure in such an environment is essential. It is for this reason, since many years, numerous articles appeared on this subject. In \cite{Taguchi}, the key issues raised by the design and the control of a piping inspection robot are presented. More generally, in the field of pipe inspections, four following major issues have to be faced: 1) how to move in a pipe; 2) how to locate in a pipe; 3) how to inspect a pipe area; 4) how to repair a default.
 Let us note that the work presented in our paper only deals with the first issue  which falls within the locomotion issue.
 To achieve locomotion in such highly constrained pipe environment, we can classify the robot designs in two categories depending whether one takes inspiration from animals living in narrow spaces or from engineering knowledge and nine associated subcategories \cite{Kassim2006a}.
 In the first category, we can imagine designs inspired from the earthworms \cite{Drapier1961}, the snakes \cite{Ikuta}, the millipedes \cite{Utsugi1979}, the Lizards \cite{Treat1997} and even from soft animals as the octopus \cite{Ginsburgh1988}.
 For the second one, we find designs based on the using of wheels and pulleys \cite{Takada1996}, the telescopic \cite{Masuda}, the impact \cite{Hyun2001} and the natural peristalsis \cite{Iddan1997}.
 The main problem of all these solutions is that each designed robot has a specific architecture for a given specification which is different in each case.
 Thus it is difficult to find the best architecture that responds to an user request.
 Despite this fact, the common denominators of all these systems are the mechanisms to adapt the contact points of the robot on the pipe surface and those to generate the expected contact forces required by the desired motion in the pipe.
 It is in this context that this paper takes place.
 Thus, the aim of the paper is to design a mechanism able to adapt the contacts on the inner surface of a pipe under constraints inherent to the environment.
 More precisely, before defining the complete architecture of our piping robot, we want  to find by an optimization  into  a mechanism selection,  the best design  minimizing the bulk volume of the robot and maximizing the transmission factor between the embedded motor and the contact points.
To simplify, in this article, we call now such a system :  the legs such as they are depicted in figure \ref{fig:robot_pipe}.
 In the context of our study, the considered pipe has a variable diameter between 28 mm and 58 mm with bends. 

The outline of the paper is as follows.
 The mechanisms that we use to make the legs of our piping inspection robot are presented in first  section. 
 Then, the objective functions and the constraints on the design parameters are introduced.
 In section three, the optimization results under the form of Pareto fronts are presented. Moreover, CAD models of candidate mechanisms are shown. Finally, the article ends with a discussion and concluding remarks.

\section*{Locomotion and environment}
The locomotion can be obtained either by putting actuated wheels or by varying the body length while the front and the rear part are alternately fixed to the pipe, the way worms do.
 These examples come from the classification made in \cite{Roh2005} where seven categories of locomotion types as (i) the pig \cite{Okamoto1999}, (ii), the wheel \cite{Choi2007}, (iii) the crawler \cite{Kwon2012}, (iv) the wall press \cite{Tsuruta2000}, (v) the walking \cite{Phee2000}, (vi) the inchworm \cite{Ono2010} and (vii) the screw type \cite{Horodinca2002}. However, it is a well-known that the locomotion can be achieved in different ways according to the type of pipe: the wheels for the straight lines and the earthworms for the bends.\\

Before presenting the retained mechanisms, let us introduce some notations. Its inner radius of the pipe $r$ varies between $r_{min}$ and $r_{max}$ with $r_{min}=14$ mm and $r_{max}=29$ mm.The minimum value is fixed in such a way that the robot can embedded a motor, electric and energy. Moreover, we define a reference frame whose the axis of motion, denoted by $x$, is along the main pipe axis and the motion of the mechanism under study is along the radius axis (Figure \ref{fig:robot_pipe}). 

\section*{Geometric variations of piping}

Park \cite{Park2011} lists a five principal geometric variations of piping. There are variations of (a) diameter, (b) curvature, and (c) inclination, (d) branched pipe, and (e) uneven inner surface. A pipe section can combine several geometric variations. For example we can have an inclined branched pipe with uneven inner surface. The variations of curvature limit the size of the robot when it negotiates the elbow \cite{Choi2007}. There are four parameters to determine the size of the robot (\ref{eq:elbow0pts}). The pipe have two  parameters with the radius of piping $r_p$ and  the curvature $r_c$. The robot is defined in a cylinder with a diameter $d_r$ and length $l_r$.
\begin{equation}
l_r=2\sqrt{(2\,r_p-d_r)(2\,r_c+d_r)}\text{ .}
\label{eq:elbow0pts}
\end{equation}


So we choice a cylinder with a diameter of $d_r=30$mm  and a length of $l_r=60$mm because the minus radius of curvature is  $r_c=45$ mm and radius of piping $r_p=20$ mm.

\section*{Selection architectures of mechanism}
We have selected two locomotion types, the inchworms .
 Then we have selected compatible  mechanisms of locomotion types and geometric variations of piping.
 The three architectures of mechanism selected is a slot-follower mechanism \cite{Valdastri2009,Kim2002,Quirini2007}, a crank and slider mechanism with 4 bars \cite{Roh2005} and 6 bars \cite{Okada1987}.
  Several mechanism have been removed. 
	the solutions of 
	Kawaguchi \cite{Kawaguchi1965} have proposed magnetic wheels but the environment is nonmagnetic. Dertien \cite{Dertien2011} have built  the Pirate robot for diameter range between 63 mm to 125 mm but the robot cannot be  miniaturized and climb vertical pipe. Suzumori \cite{Suzumori1999} have designed a micro inspection robot for recover an object in a 25 mm pipe, but drive system \cite{Miyagawa2007} is limited a small pipe diameter variation. 
Anthierens's thesis \cite{Anthierens,Anthierens2000a} focuses on an inchworms locomotion in steam generator but variations and  curvatures of  diameter are not possible.



\section*{A slot-follower mechanism, a crank and slider mechanism with 4 bars and 6 bars}
From the literature review, we have selected three architectures of mechanism suitable to be used according to the size of the pipe. 
 For the three mechanisms, $O$ is the origin of the reference frame, $P(x,y)$ is the coordinate of the end-effector and $\rho$ is the length of the prismatic joint $OA$. 
\begin{figure}[htb!]
	\centering	
		\includegraphics[scale=0.25]{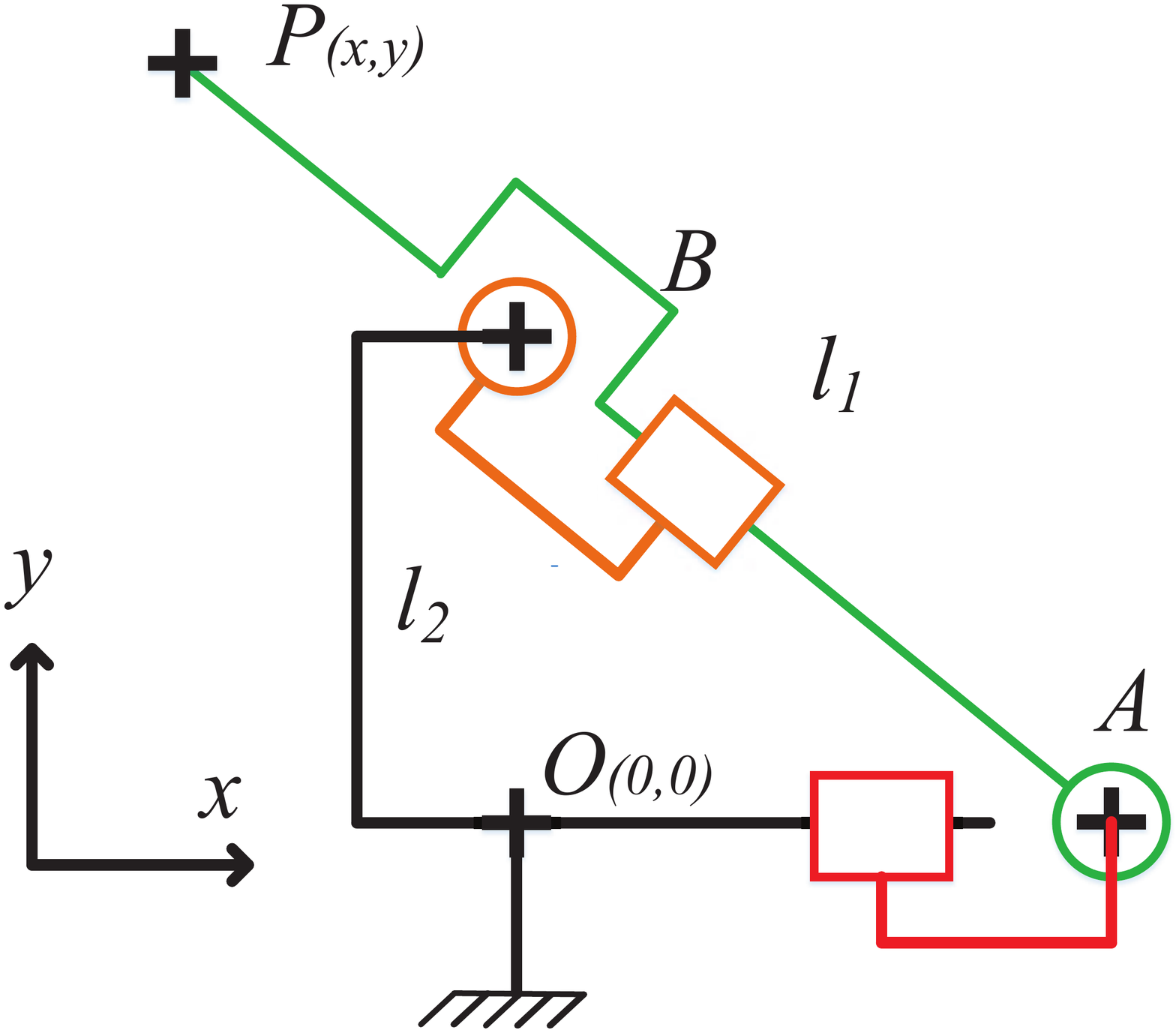}
	\caption{A slot-follower mechanism.}
	\label{fig:biellette-schemat}
\end{figure}

Figure \ref{fig:biellette-schemat} is a slot-follower mechanism \cite{Valdastri2009,Kim2002,Quirini2007} where the fixed prismatic joint is actuated and the other joints are idle. Lengths $l_1$ and $l_2$ denoting the lengths of $AP$ and $OB$ respectively define the geometry of mechanism entirely. The serial singularities (working mode changing) occur when the points $A$ and $O$ are coincided. The mechanical limits occur when the points $P$ and $B$ are coincided.
\begin{figure}[htb!]
	\centering
		\includegraphics[scale=0.20]{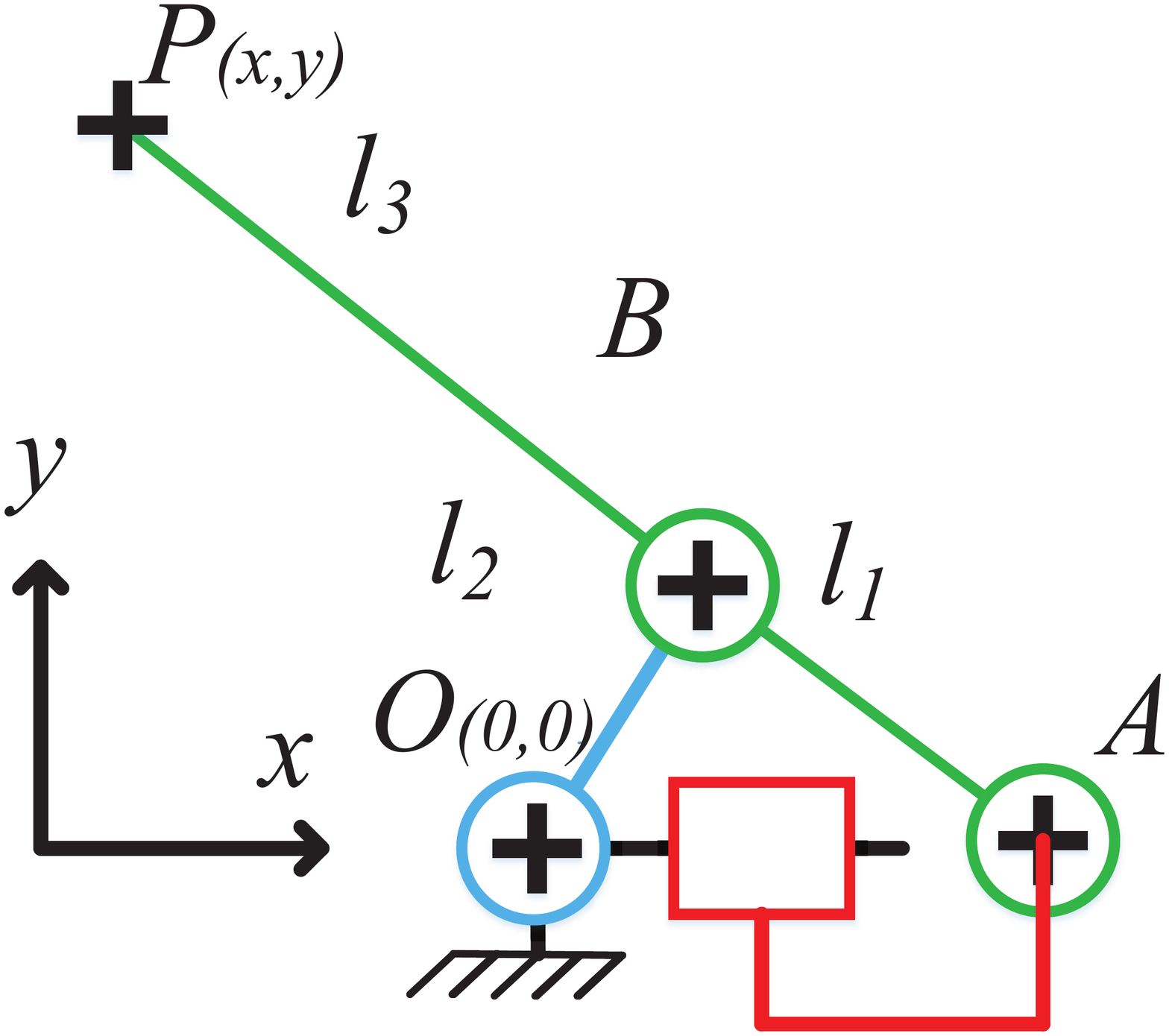}
	\caption{Crank and slider mechanism with 4 bars.}
	\label{fig:3barres-schemat}
\end{figure}

Figure \ref{fig:3barres-schemat} is a crank and slider mechanism with 4 bars \cite{Roh2005} where the fixed prismatic joint is actuated and the other joints are idle. Lengths $l_1$, $l_2$ and $l_3$ denoting the lengths of $AB$, $OB$ and $BP$ respectively define the geometry of mechanism entirely. The parallel singularities occur when $O$, $B$ and $A$ are aligned.
\begin{figure}[!htbp]
	\centering
		\includegraphics[scale=0.25]{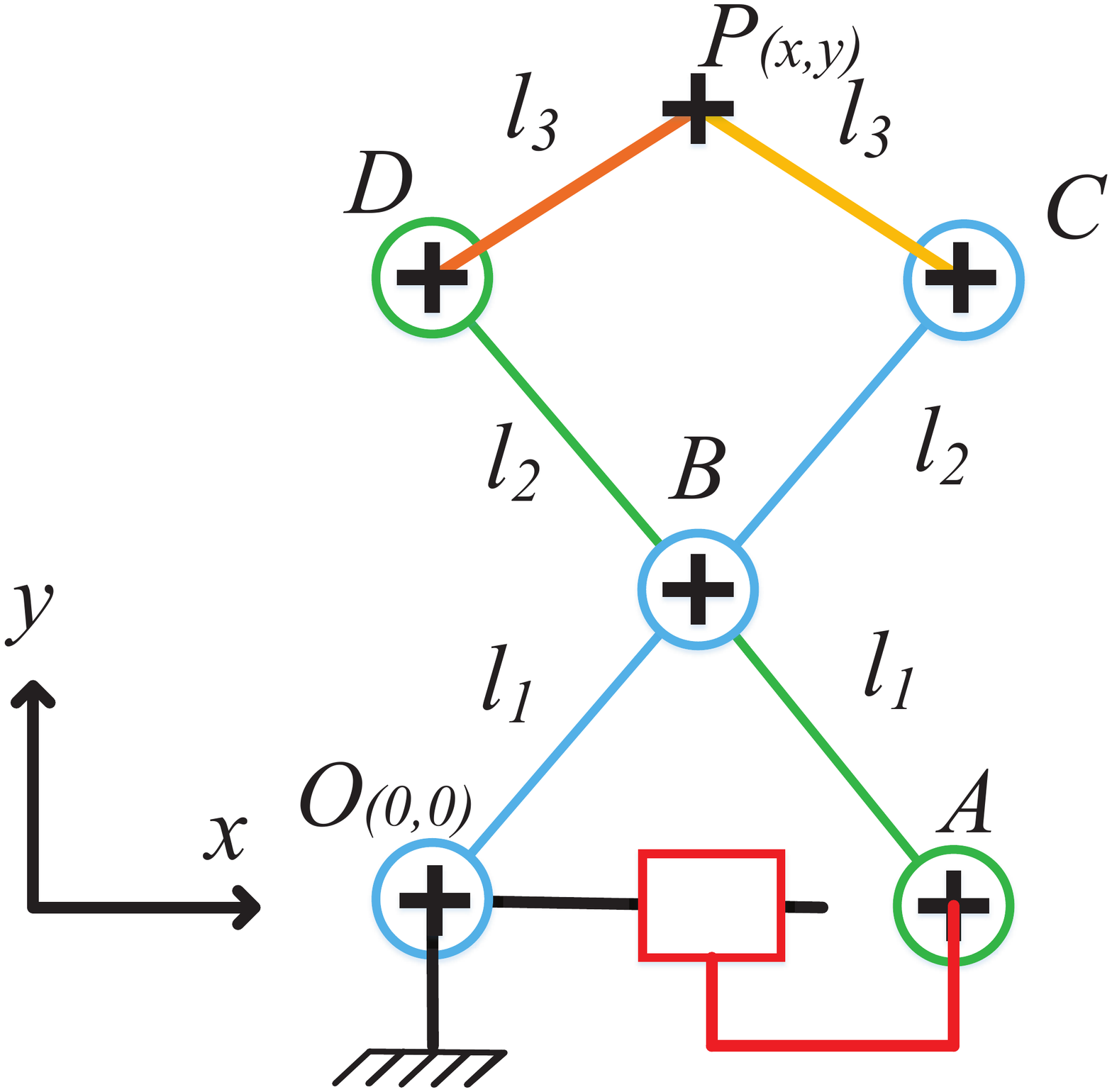}
	\caption{Crank and slider mechanism with 6 bars.}
	\label{fig:6barres-schemat}
\end{figure}

Figure \ref{fig:6barres-schemat} is a crank and slider mechanism with 6 bars \cite{Okada1987} where one fixed prismatic joint is actuated and the other joints are idle. To reduce the design parameter space, we have added some equalities. Lengths $l_1$ denotes the lengths of $AB$ and $OB$, lengths $l_2$ denotes the lengths of $BD$ and $BC$ and lengths $l_3$ denotes the lengths of $DP$ and $CP$. These simplifications will be justified by the conclusion of the optimization of the crank and slider mechanism with 4 bars (See Section V). The parallel singularities occur when $O$, $B$ and $A$ are aligned or $C$, $D$ and $P$ are aligned.

\subsection*{Kinematic modeling of three mechanisms}
\begin{figure}[htb!]
	\centering
		\includegraphics[scale=0.25]{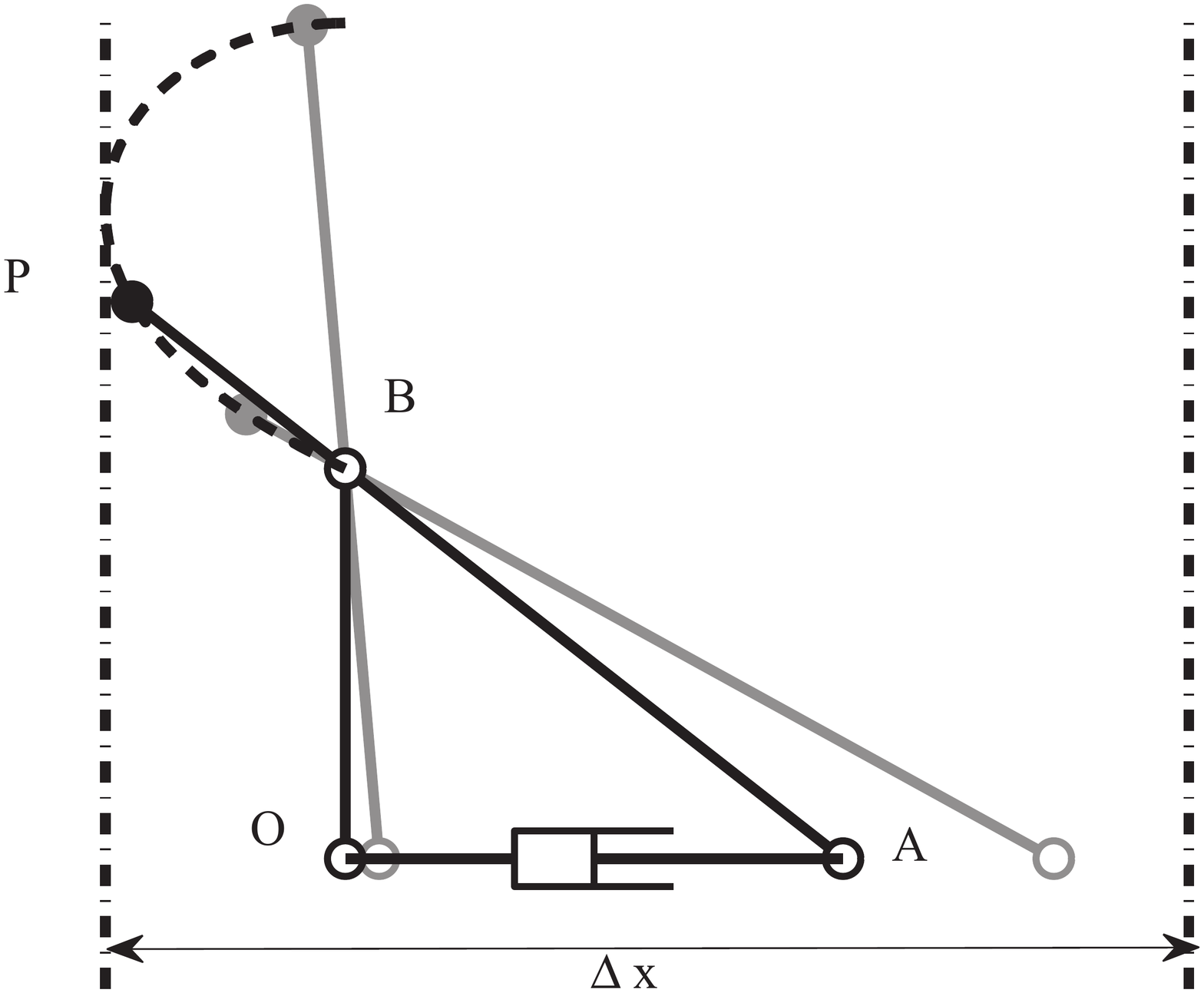}
	\caption{The black lines represent a slot-follower mechanism at an intermediate configuration. The gray lines correspond to the lower and upper configurations of the mechanism. The dotted line is the trajectory of $P$ and $\Delta x$ is the size of the mechanism.}
	\label{fig:biellet-schemat-traje}
\end{figure}

\begin{figure}[htb!]
	\centering
		\includegraphics[scale=0.25]{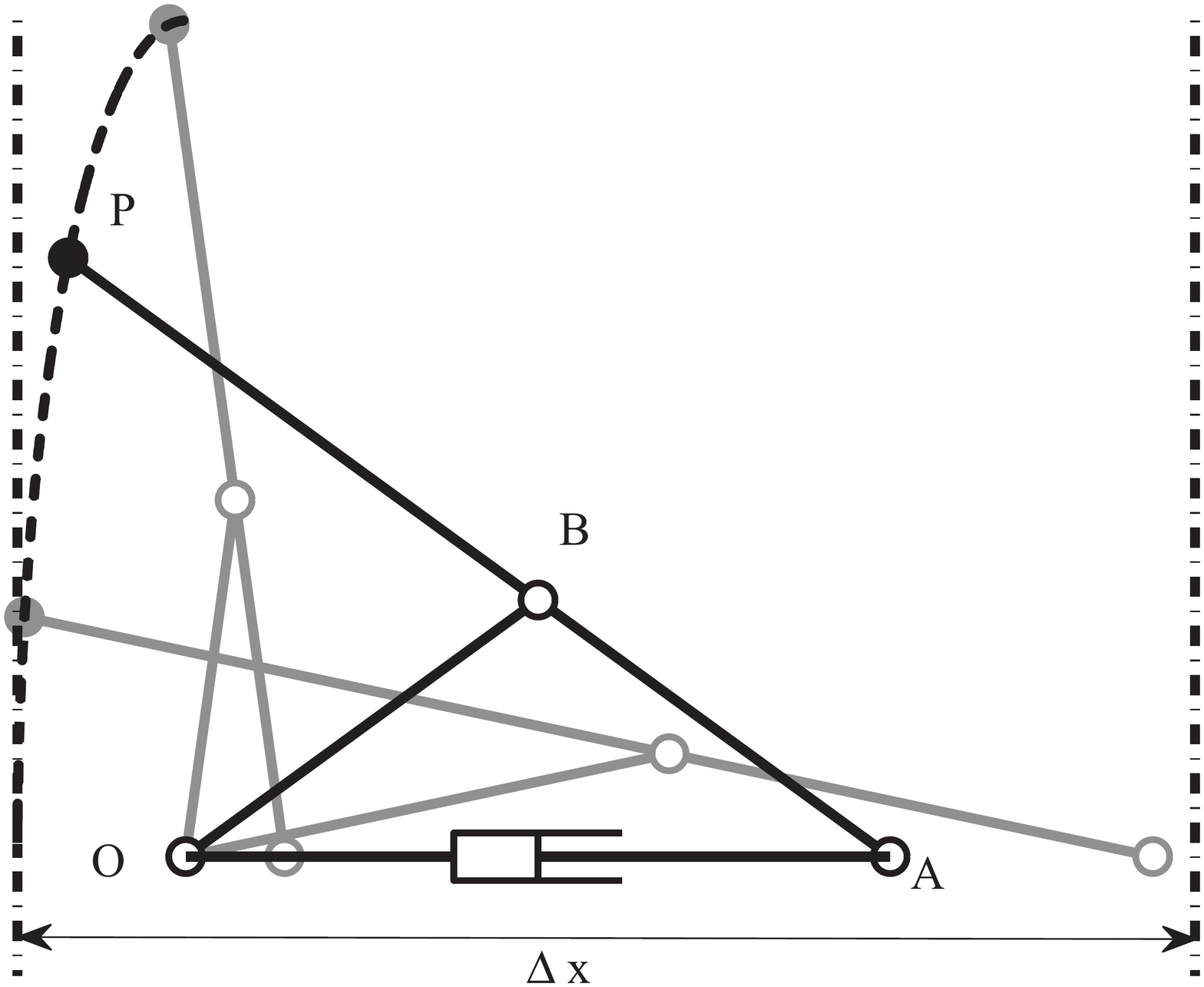}
	\caption{The black lines represent a Crank and slider mechanism with 4 bars at an intermediate configuration. The gray lines correspond to the lower and upper configurations of the mechanism. The dotted line is the trajectory of $P$ and $\Delta x$ is the size of the mechanism.}
	\label{fig:3barres-schemat-traje}
\end{figure}

\begin{figure}[htb!]
	\centering
		\includegraphics[scale=0.25]{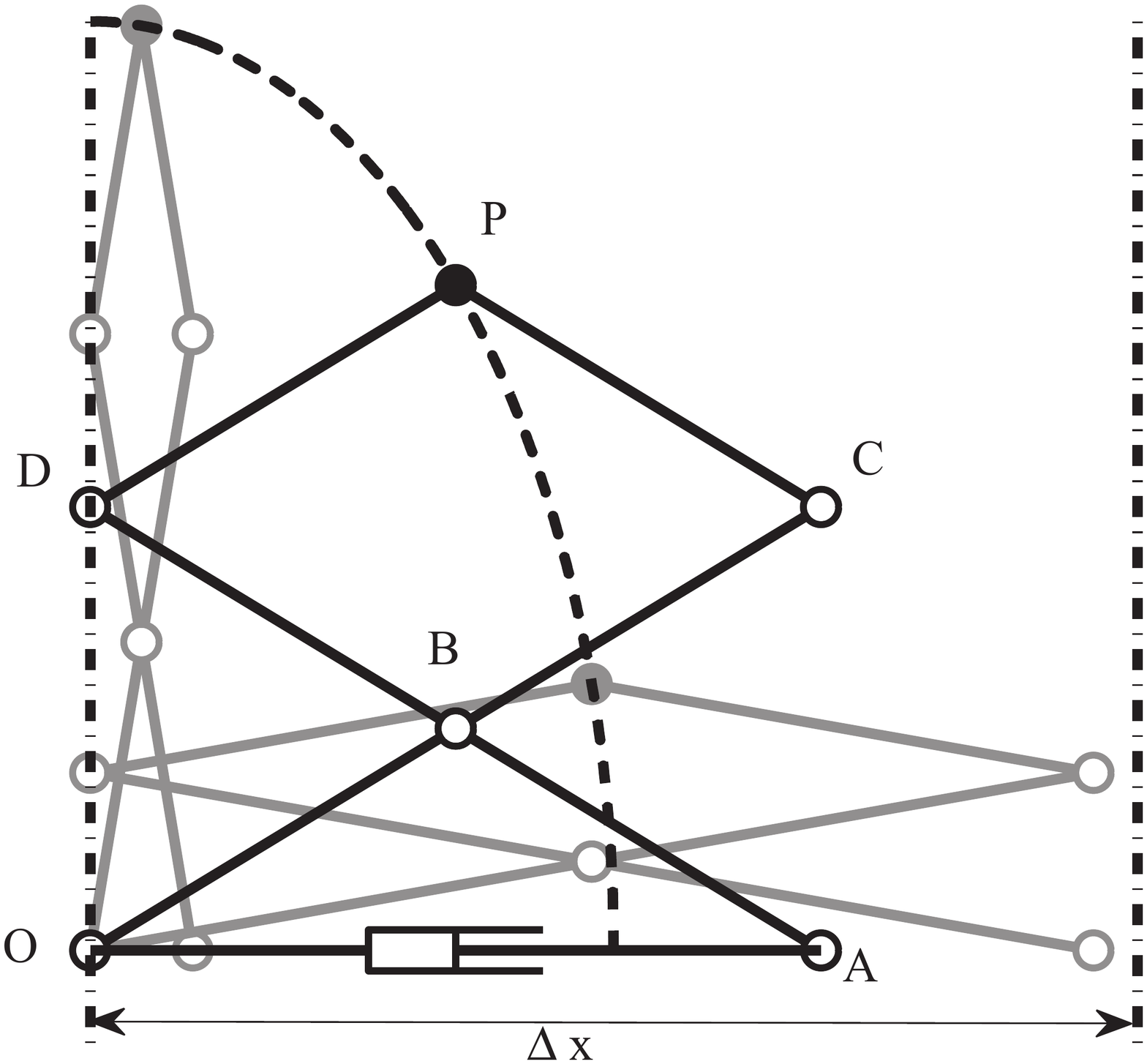}
	\caption{The black lines represent a Crank and slider mechanism with 6 bars at an intermediate configuration. The gray lines correspond to the lower and upper configurations of the mechanism. The dotted line is the trajectory of $P$ and $\Delta x$ is the size of the mechanism.}
	\label{fig:6barres-schemat-traje}
\end{figure}

The geometric parameters of the mechanism, $l_1$, $l_2$ , $l_3$ (only 4 bars and 6 bars mechanism)  and the actuator displacements $\rho$ permit us to define the Direct Kinematics Model (DKM) to have the relation between the geometric parameters of the mechanism, the actuator displacements and the moving platform pose $y$.
%
%
The assembly mode is chosen such that $y>0$ and $y>y_B$, where $y_B$ is the ordinate coordinate of the point $B$. We also add constraints to avoid the singular configurations. The limits of $y$ is defined by $y_{max}$ and $y_{min}$.
%
%
Figures \ref{fig:biellet-schemat-traje}--\ref{fig:6barres-schemat-traje} depict the three mechanisms in their lower, upper and intermediate configurations for the assembly mode chosen in the paper. The size of the mechanism is defined by $\Delta x$.

For a {\it slot-follower mechanism} (Figure \ref{fig:biellet-schemat-traje}), a single assembly mode exists and we have:
\begin{equation}
 y={\frac {{l_2}\,{l_1}}{\sqrt {{{l_2}}^{2}+{\rho}^{2}}}} \text{ .} \label{MGD_1}
\end{equation}
The limits $y_{max}$ and $y_{min}$ are:
\begin{eqnarray}
 y_{min}=l_2\text{ and } y_{max}=l_1\text{ .} \label{MGD_1_2}
\end{eqnarray}
$y_{min}$ is constrained by the mechanical limits, when the points $P$ and $B$ are coincide and $y_{max}$ is constrained by the length of $l_1$ in vertical, when the points $A$ and $O$ are coincide (Figure \ref{fig:biellet-schemat-traje}).
As for a {\it crank and slider mechanism with 4 bars} (Figure \ref{fig:3barres-schemat-traje}), two assembly modes exist and we set
\begin{align}
 y=\frac {\sqrt {2 l_1^2 l_2^2-l_1^4+2 l_1^2\rho^2-l_2^4+2\rho^2 l_2^2-\rho^4} \left( l_2+ l_3 \right) }{2\rho l_2} \text{ .} \label{MGD_2}
\end{align}
The limits $y_{max}$ and $y_{min}$ are:
\begin{align}
 y_{min}=0\text{ and }  y_{max}=min \left ( l_2\left ( \frac{l_3}{l_1}+1 \right),l_1+l_3 \right) \text{ .} \label{MGD_2_2}
\end{align}
$y_{min}$ is zero when the  mechanism  is in parallel singularities because the points $O$, $B$ and $A$ are aligned.

As for a {\it crank and slider mechanism with 6 bars} (Figure \ref{fig:6barres-schemat-traje}), four assembly modes exist and we set
\begin{equation}
y={\frac {\sqrt {a_1+2\,\sqrt {a_2}}}{{ 2 l_1}}}\text{ ,} \label{MGD_3}
\end{equation}
with:
\begin{align}
a_1=& 4\,{{ l_1}}^{4}+8\,{{ l_1}}^{3}{ l_2}+4\,{{ l_1}}^{2}{{ l_2}}^{2}+4\,{{ l_1}}^{2}{{ l_3}}^{2}-{{ l_1}}^{2}{\rho}^{2} \notag \\
 &-2\,{ l_1}\,{ l_2}\,{\rho}^{2} -2\,{{ l_2}}^{2}{\rho}^{2} \text{ ,}\notag \\
a_2=& \left(2\,{ l_1}-\rho \right) \left( 2\,{ l_1}+\rho \right) \left( { l_1}+{ l_2} \right) ^{2} \notag \\
 & \left( 2\,{ l_1}\,{ l_3}-{ l_2}\,\rho \right) \left( 2\,{ l_1}\,{ l_3}+{ l_2}\,\rho \right) \text{ .}\notag
\end{align}
The limits $y_{max}$ an $y_{min}$ are:
\begin{align}
 & y_{max}=l_1+l_2+l_3 \text{ ,} \\
	& {\rm if} \quad l_3 \geq l_2 \quad {\rm then } \quad y_{min}=\sqrt{(l_3)^2-(l_2)^2}\text{ ,} \\
 & {\rm else} \quad y_{min}=\sqrt{(l_2)^2-(l_3)^2} \left ( \frac{l_1}{l_2}+1\right ) \text{ .}	\label{MGD_3_2}	
\end{align}
$y_{max}$ is equal $l_1+l_2+l_3$ because the  mechanism  is in serial singularities when the all points are aligned in vertical. $y_{min}$ is constrain by parallel singularities.

\section*{MULTI-OBJECTIVE DESIGN OPTIMIZATION}
\subsection*{Objective functions}
The multi-objective optimization problem aims to determine the optimum geometric parameters of  the leg mechanism in order to minimize the size of the mechanism and to maximize the transmission factor.
\subsubsection*{Size of the mechanism}
There are numerous ways to define the compactness of a mechanism. 
The first objection function is
\begin{equation}
f_1= \Delta x\text{ .}
\end{equation}
where $\Delta x$ is the size of ​​the mechanism on the $x$ axis which is directly connected with the swept volume of the mechanism. $\Delta x$ is obtained by the projection onto the $x$-axis of any points of the mechanism during its motion. We set: 
\begin{equation}
	\rho \geq 0.5 {\rm ~mm} \text{ and } \Delta x \leq 35 {\rm ~mm}.
\end{equation}
The first limit avoids the serial singularities \cite{chablat1998} and the second one restricts the mechanism size. 
\subsubsection*{Transmission force efficiency}
The second objective function is the transmission force efficiency which is defined as the ratio between output and input forces. We note
\begin{equation}
f_2=\eta_f=\frac{F_w}{F_a}\text{ .}
\end{equation}
where $F_a$ is the actuator force along the $x$ axis, $F_w$ is the contact force in $P$ along the $y$ axis and $\eta_f$ is the transmission efficiency between the two forces. 
Moreover, for convenience, we set the transmission efficiency as follows: 
\begin{equation}
\eta_f \geq 0.3\text{ .}
\end{equation}
In accordance with the definition of the transmission efficiency, for a {\it slot-follower mechanism}, we have: 
\begin{equation}
\eta_f={\frac { \left( {{ l_2}}^{2}+{\rho}^{2} \right) ^{3/2}}{{ l_2}\,{ l_1}\,\rho}}\text{ .}
\end{equation}
As regarding the {\it crank and slider mechanism with 4 bars}, we have: 
\begin{equation}
\eta_f=2\,{\frac {\sqrt {-{l_1}^{4}+2\,{l_1}^{2}{l_2}^{2}+2\,{l_1}^{2}{\rho}^{2}-{l_2}^{4}+2\,{l_2}^{2}{\rho}^{2}-{\rho}^{4}}{\rho}^{2}l_2}{ \left( l_2+l_3 \right) \left( {l_1}^{2}-{l_2}^{2}+{\rho}^{2} \right) \left( {{\it L1}}^{2}-{l_2}^{2}-{\rho}^{2} \right) }} \text{ .}
\end{equation}
Finally, for a {\it crank and slider mechanism with 6 bars}, we have: 
\begin{equation}
	\eta_f=2\,{\frac {l_1\,\sqrt {B}\sqrt {A}}{\rho\,C}} \text{ ,}
\end{equation}
with
\begin{align}
	A&=\left( 2\,l_1-\rho \right) \left( 2\,l_1+\rho \right) \left( l_1+l_2 \right) ^{2} \left( 2\,l_3\,l_1-l_2\,\rho \right) \notag \\
	&\,\,\, \left( 2\,l_3\,l_1+l_2\,\rho \right) \text{ ,} \\
	B&=4\,{l_1}^{4}+8\,{l_1}^{3}l_2+4\,{l_1}^{2}{l_2 }^{2}-2\,{l_2}^{2}{\rho}^{2}+4\,{l_3}^{2}{l_1}^{2} \notag\\
	& \,\,\,-{l_1}^{2}{\rho}^{2}-2\,l_1\,{\rho}^{2}l_2+2\,\sqrt {A} \text{ ,} \\
	C&= 2\,{l_2}^{2}\sqrt {A}+{l_1}^{2}\sqrt {A}+2\,l_1\,l_2\,\sqrt {A}+4\,{l_2}^{4}{l_1}^{2} \notag\\
	&\,\,\,+8\,{l_2}^{3}{l_1}^{3}-2\,{l_2}^{4}{\rho}^{2}-2\,{l_1}^{2}{l_2}^{2}{\rho}^{2}- \notag\\
	&\,\,\,4\,{l_2}^{3}{\rho}^{2}l_1+4\,{l_2}^{2}{l_1}^{4}+4\,{l_3}^{2}{l_1}^{4} \notag\\
	&\,\,\,+8\,l_2\,{l_3}^{2}{l_1}^{3}+4\,{l_1}^{2}{l_2}^{2}{l_3}^{2} \text{ .} \notag
\end{align}
\subsection*{Design constraints}
Due to assembly constraints, we have a set of inequalities. For a {\it slot-follower mechanism}, we have:
\begin{align}
	& l_1 \geq r_{max}\text{ and } l_2 \leq r_{min}\text{ .} 
\end{align}
For a {\it crank and slider mechanism with 4 bars}, we have: 
\begin{align}
	&min \left ( l_2\left ( \frac{l_3}{l_1}+1 \right),l_1+l_3 \right) \geq r_{max} \text{ and }r_{min}= 0. 
\end{align}
For a {\it crank and slider mechanism with 6 bars}, we have: 
\begin{align}
	& l_1+l_2+l_3 \geq r_{max}\text{ ,} \\
 & {\rm if} \quad l_2 \ge l_3 \quad {\rm then } \quad \frac{\sqrt{{l_2}^2-{l_3}^2} }{l_2} \left(l_1+l_2 \right) \leq r_{min}\text{ ,}  \\
	& {\rm else} \quad 	\sqrt{{l_3}^2-{l_2}^2 } \leq r_{min}\text{ .}  
\end{align}
\subsection*{Design Variables}
Along with the above mentioned geometric parameters $l_1$, $l_2$ and $l_3$ are considered as
design variables, also called decision variables. 
As there are three leg mechanisms under study, the leg type is another design variable that has to be taken into account. Let $d$ denote the leg type: $d = 1$ stands for the slot-follower mechanism; $d = 2$ stands for crank and slider mechanism with 4 bars and $d = 3$ stands for the crank and slider mechanism with 6 bars.
As a result, the optimization problem contains one discrete variable, $i.e.$, $d$, and three continuous design variables, $i.e.$, $l_1$, $l_2$, $l_3$. Hence, the design variables vector $\bf x$ is given by:
\begin{equation}
 {\bf{x}} = \left[ d, l_1, l_2, l_3 \right]^T
\end{equation}
\subsection*{Multi-objective optimization problem statement}
The Multi-objective Optimization Problem (MOO) for a leg mechanism can be stated as:
{\it Find the optimum design variables $\bf x$ of leg mechanism in order to minimize the size of the mechanism and maximize the transmission factor subject to geometric constraints.} Mathematically, the problem can be written as:
\begin{equation}
\label{Eq:MOO_PRR} 
\left\{ \begin{array}{l}
{\rm minimize~} f_1(\mathbf{x})= \Delta x\text{ ;} \notag \\
{\rm maximize~} f_2(\mathbf{x})= \eta_f\text{ .} \notag
\end{array} \right. \notag \\
\end{equation} 
\begin{align}
{\rm over} & \quad {\bf x}=\left[ d, l_1, l_2, l_3 \right]^T \notag \\
& \notag \\
{\rm subject~to:} &\notag \\
	g_1:& \quad l_1 \geq r_{max} {\rm ~for~} d=1 \text{ ;}\notag \\
	g_2:& \quad l_2 \leq r_{min} {\rm ~for~} d=1 \text{ ;}\notag \\
	g_3:& \quad min \left ( l_2\left ( \frac{l_3}{l_1}+1 \right),l_1+l_3 \right) \geq r_{max} {\rm~for~} d=2 \text{ ;}\notag \\
	g_4:&\quad r_{min} \geq 0 {\rm~for~} d=2 \text{ ;}\notag \\
	g_5:& \quad l_1+l_2+l_3 \geq r_{max} {\rm~for~} d=3 \text{ ;}\notag \\
 g_6:& \quad {\rm if } \quad l_2 \ge l_3 \quad {\rm then } \quad \frac{\sqrt{{l_2}^2-{l_3}^2} }{l_2} \left(l_1+l_2 \right) \leq r_{min} \notag \\
				& \quad {\rm else} \quad 	\sqrt{{l_3}^2-{l_2}^2 } \leq r_{min} {\rm~for~} d=3 \text{ ;}\notag \\
 g_7:&\quad 	\eta_f \geq 0.3 \text{ ;}\notag \\
	g_8:&\quad 	\rho \geq 0.5 {\rm ~mm} \text{ ;}\notag \\
	g_9:&\quad 	\Delta x \leq 35 {\rm ~mm} \text{ ;}\notag \\
	g_{10}:&\quad 	1 {\rm ~mm} \leq l_1, l_2, l_3 \leq 50 {\rm ~mm} \text{ .}\notag 	
\end{align}
\subsection*{Optimization implementation}
The classic approach to optimization is to calculate all possible combinations in the solution space. With five settings ($l_1$,$l_2$,$l_3$,$d$,$\rho$), the computation time becomes excessive. For example, we have 12 billion combinations with 250 values for $l_1$,$l_2$,$l_3$,$\rho$ and 3 values of $d$.

To solve the optimization problem, we use the genetic algorithm of Matlab \cite{chablat2010}. A genetic algorithm is a search heuristic that mimics the process of natural selection. The computation time per mechanism is 2 hours with the settings:
\begin{itemize}
	\item population size: $6000$;
	\item pareto fraction: $50\%$;
	\item tolerance function: $10^{-4}$;
	\item number of sessions per problem: $5$.
\end{itemize}
To scan the solution space, the algorithm uses a population of $6000$ individuals with $50\%$ of individuals on the Pareto front.
This allows the other half of the population to reach other optimal solutions.
For the same reason, we execute five sessions per mechanism in order to have redundancy.
To reduce the computation time, tolerance precision of the objectives functions is $10^{-4}$.
The computer use an Intel Core i5-560 at 2.67GHz with 8 Go RAM.
\section*{RESULTS AND DISCUSSIONS}
\begin{figure}[htb!]
	\centering
		\includegraphics[width=8cm]{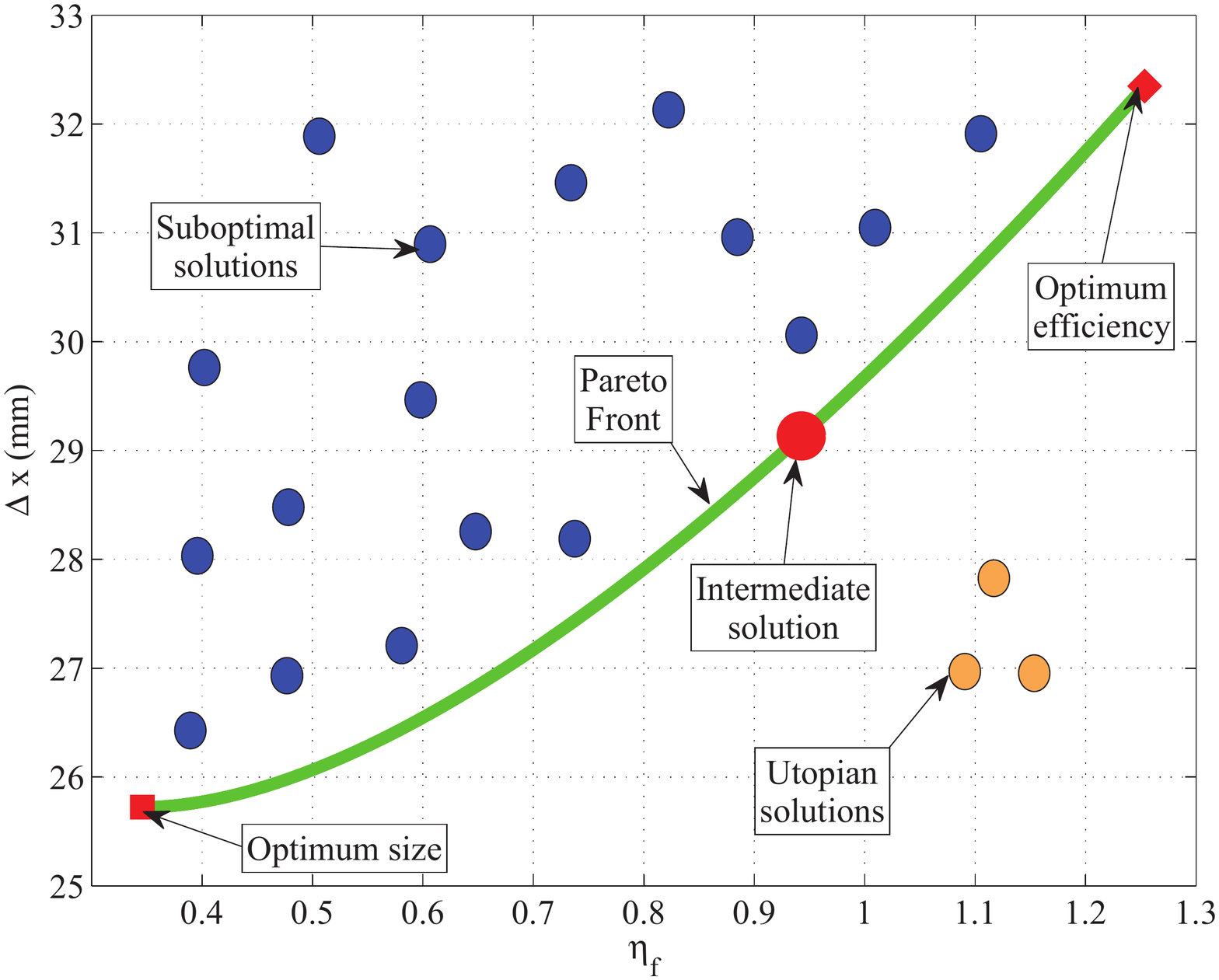}
	\caption{Example of the Pareto front with a mechanism versus the two objective functions $ \Delta x$ and $\eta_f$. The Pareto front includes optimal solutions. The suboptimal solutions are feasible but inferior to the Pareto front. Utopian solutions are to better than optimal solutions but infeasible.}
	\label{fig:Solution}
\end{figure}

Based on the kinematic models introduced in section III, the multi-objective optimization problem, expressed in section IV, is solved by means MAPLE and MATLAB. MAPLE codes given the expression of the objective functions and MATLAB code provides functions for the optimization and to make filter to obtain the Pareto front. Figure \ref{fig:Solution} depicts the Pareto front obtained after the optimization. 

\begin{figure}[htb!]
	\centering
		\includegraphics[width=8.5cm]{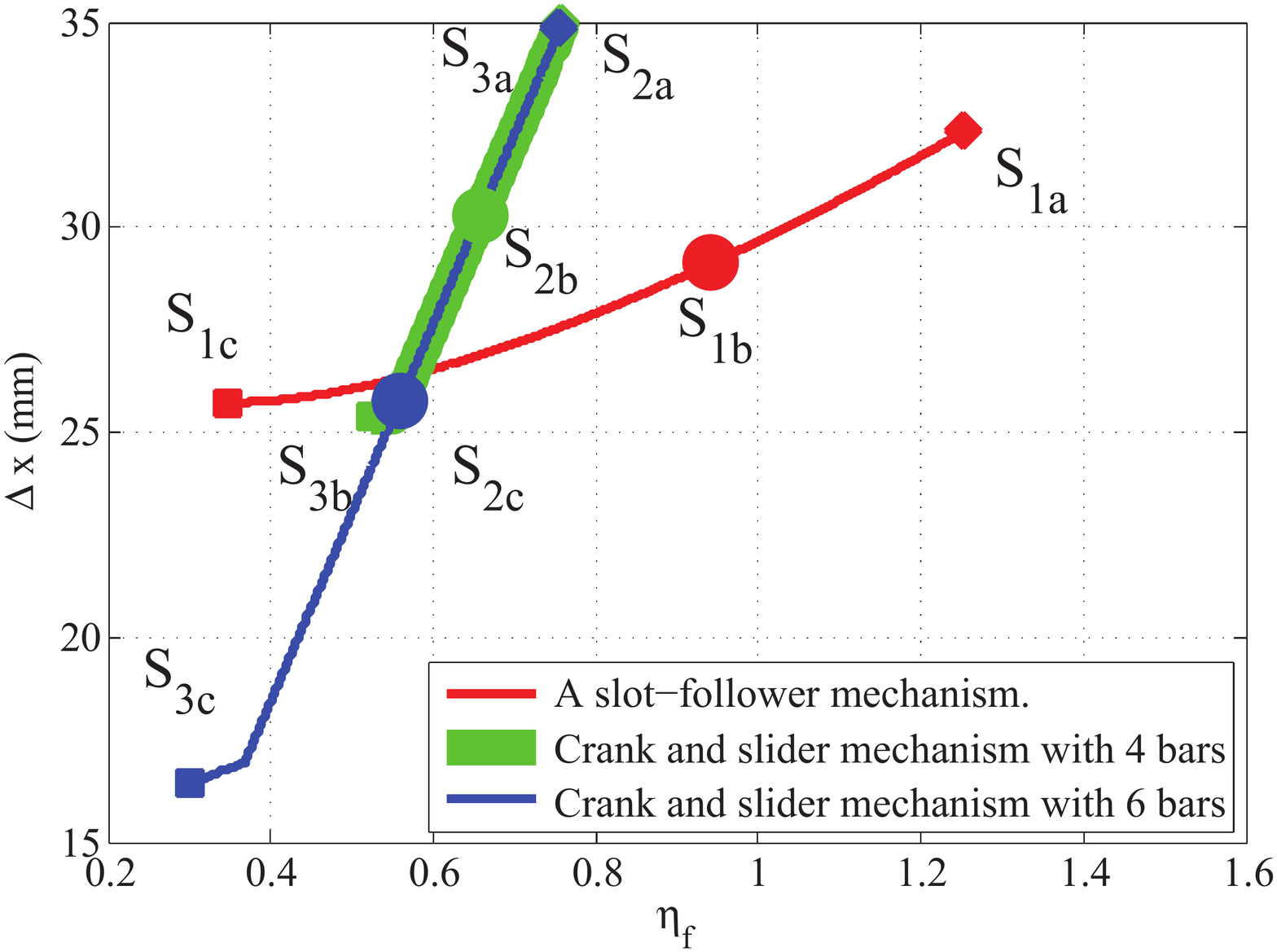}
	\caption{Pareto front associated with the three mechanisms with the two objective functions $\Delta x$ and $\eta_f$.}
	\label{fig:All_Solutions}
\end{figure}

To allow a better understanding of the Pareto front, we made three optimizations by setting d = 1, 2, 3, $i.e.$ the type of mechanism, as is depicted in Figure~\ref{fig:All_Solutions}. 

\subsection*{Slot-follower mechanism}
\begin{figure}[htb!]
	\centering
		\includegraphics[width=8.5cm]{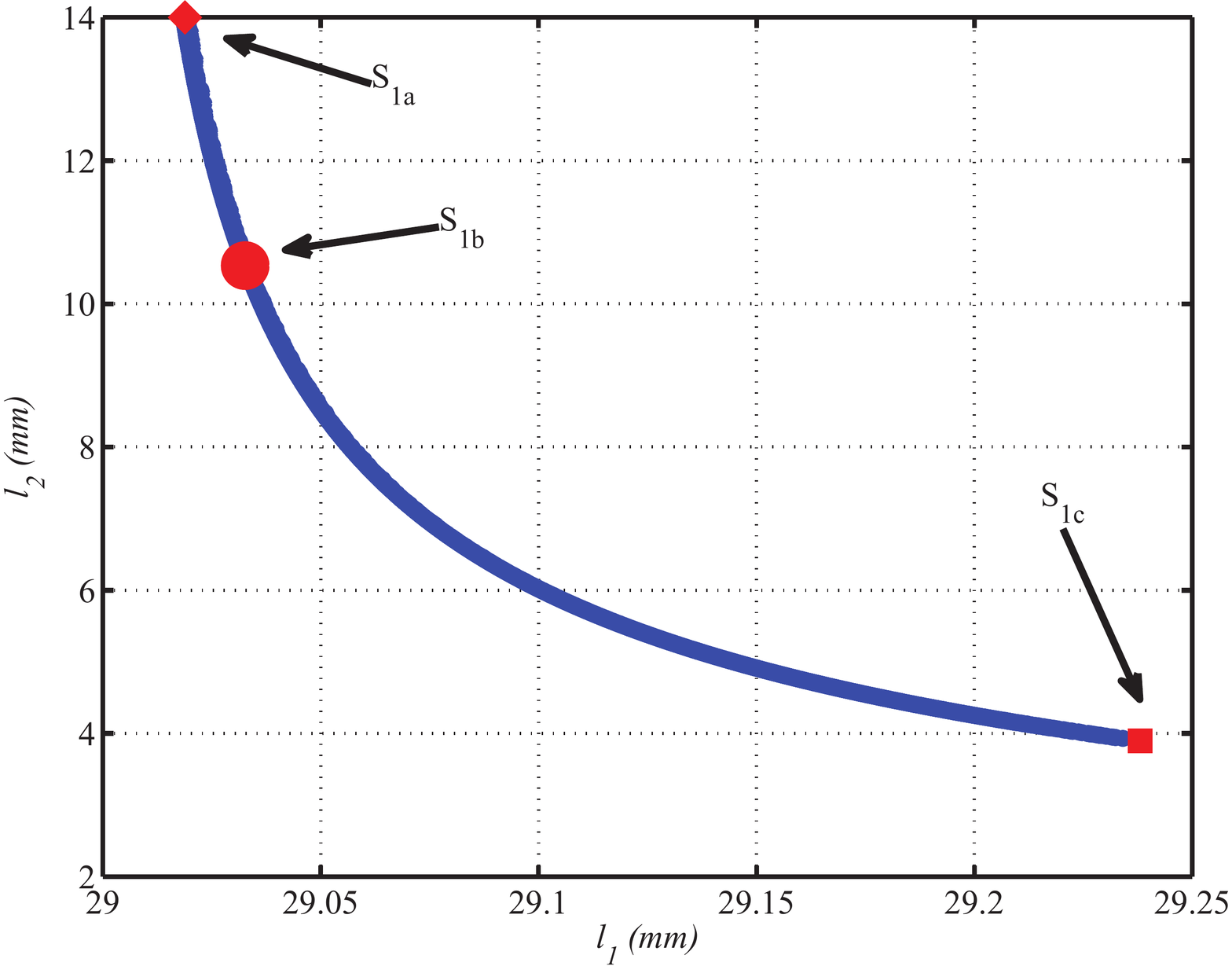}
	\caption{Pareto front for the slot-follower mechanism in the design parameter space. $S_{1a}$ is an optimum efficiency solution, $S_{1b}$ is an intermediate solution and $S_{1c}$ is an optimum size solution.}
	\label{fig:Solution_Bielle}
\end{figure}

\begin{figure}[htb!]
	\centering
		\includegraphics[scale=0.70]{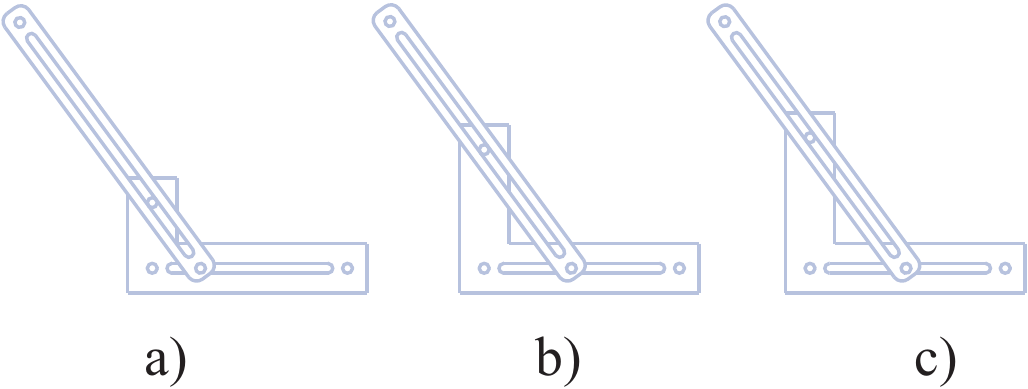}
	\caption{CAD model of the three solutions from the Pareto front for the slot-follower mechanism with a) the optimum size solution, b) the intermediate solution, c) the optimum efficiency solution.}
	\label{fig:CAD_Bielle}
\end{figure}

\begin{figure}[htb!]
	\centering
		\includegraphics[width=8.5cm]{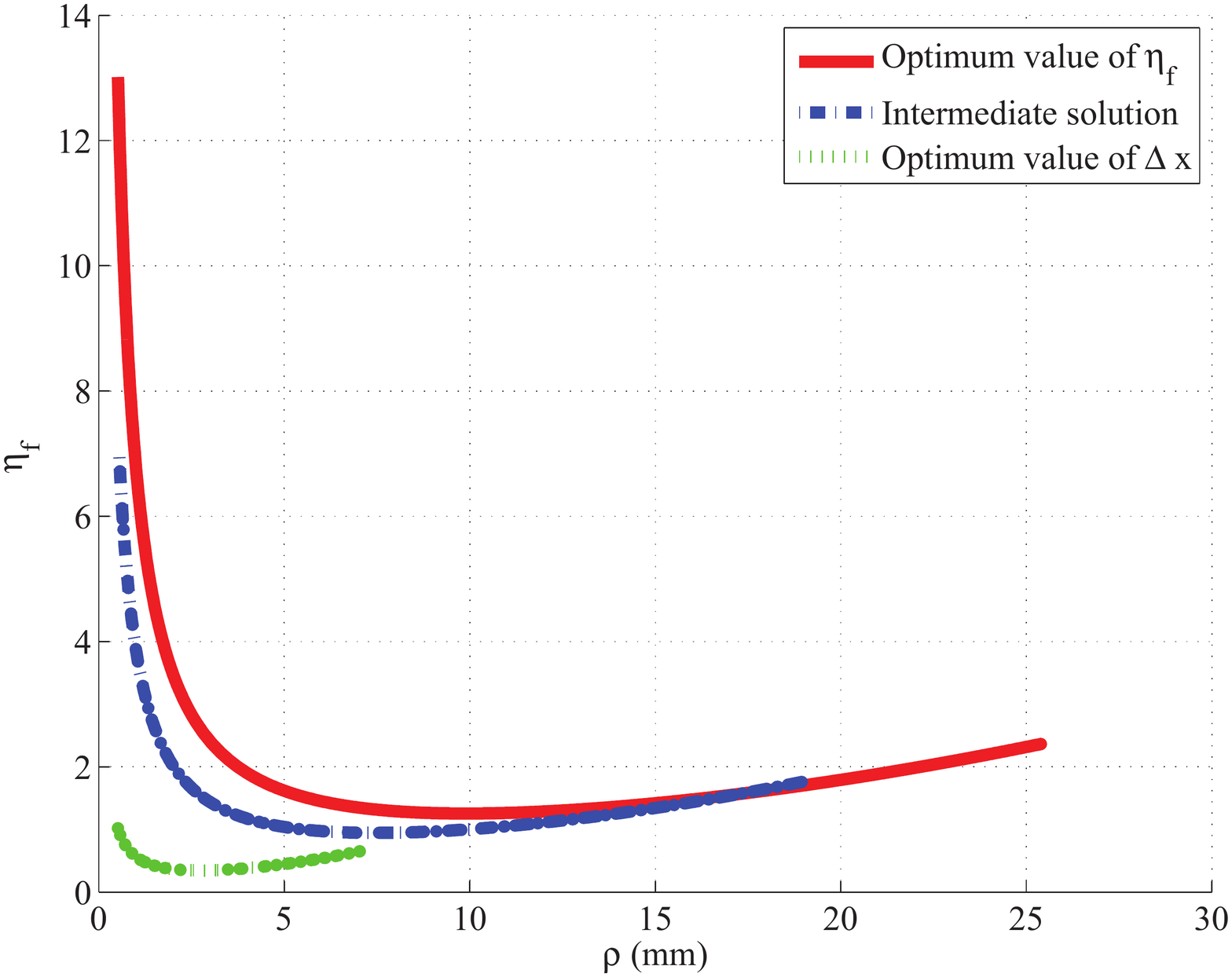}
	\caption{Evolution of the transmission force efficiency between the high and low position of the slot-follower mechanism.}
	\label{fig:Trace_Bielle}
\end{figure}

From the Pareto front, we can extract three values:
\begin{itemize}
 \item optimum value of $\Delta x = 25.7 {\rm ~mm}$ with $\eta_f =0.35$:
\end{itemize}
 \begin{equation}
l_1=29.2 {\rm ~mm}\text{ and } l_2=3.9 {\rm ~mm} \text{ ;}
 \end{equation}
\begin{itemize}
 \item optimum value of $\eta_f =1.25 $ with $\Delta x =32.3$: 
\end{itemize}
 \begin{equation}
 l_1=29.0 {\rm ~mm}\text{ and } l_2=14.0 {\rm ~mm}\text{ ;}
 \end{equation}
\begin{itemize}
 \item intermediate solution $\Delta x =29.1$ and $\eta_f =0.94$: 
\end{itemize}
 \begin{equation}
 l_1=29.0 {\rm ~mm}\text{ and } l_2=10.5 {\rm ~mm}\text{ .}
 \end{equation}
Figure \ref{fig:Solution_Bielle} depicts the evolution of the design variables and Figure~\ref{fig:CAD_Bielle} the CAD model associated with the optimal value of $\Delta x$ and $\eta_f$ and an intermediate solution. We note that $l_2=r_{min}$.
Figure \ref{fig:Trace_Bielle} depicts the evolution of the transmission force efficiency between the lower and upper configurations of the slot-follower mechanism. 
The constraint $g_8$ prevents a serial singularity. So the lever arm between points $P$, $B$ and $C$ intervenes. 
The value of transmission force efficiency is high for the optimum efficiency and Intermediate solutions because the  length between $P$ and  $B$ is small with  $l_2 \approx r_{min}$. In contrast to the optimum size  with a  $l_2$ small. 
\subsection*{Crank and slider mechanism with 4 bars}
\begin{figure}[htb!]
	\centering
		\includegraphics[width=8.5cm]{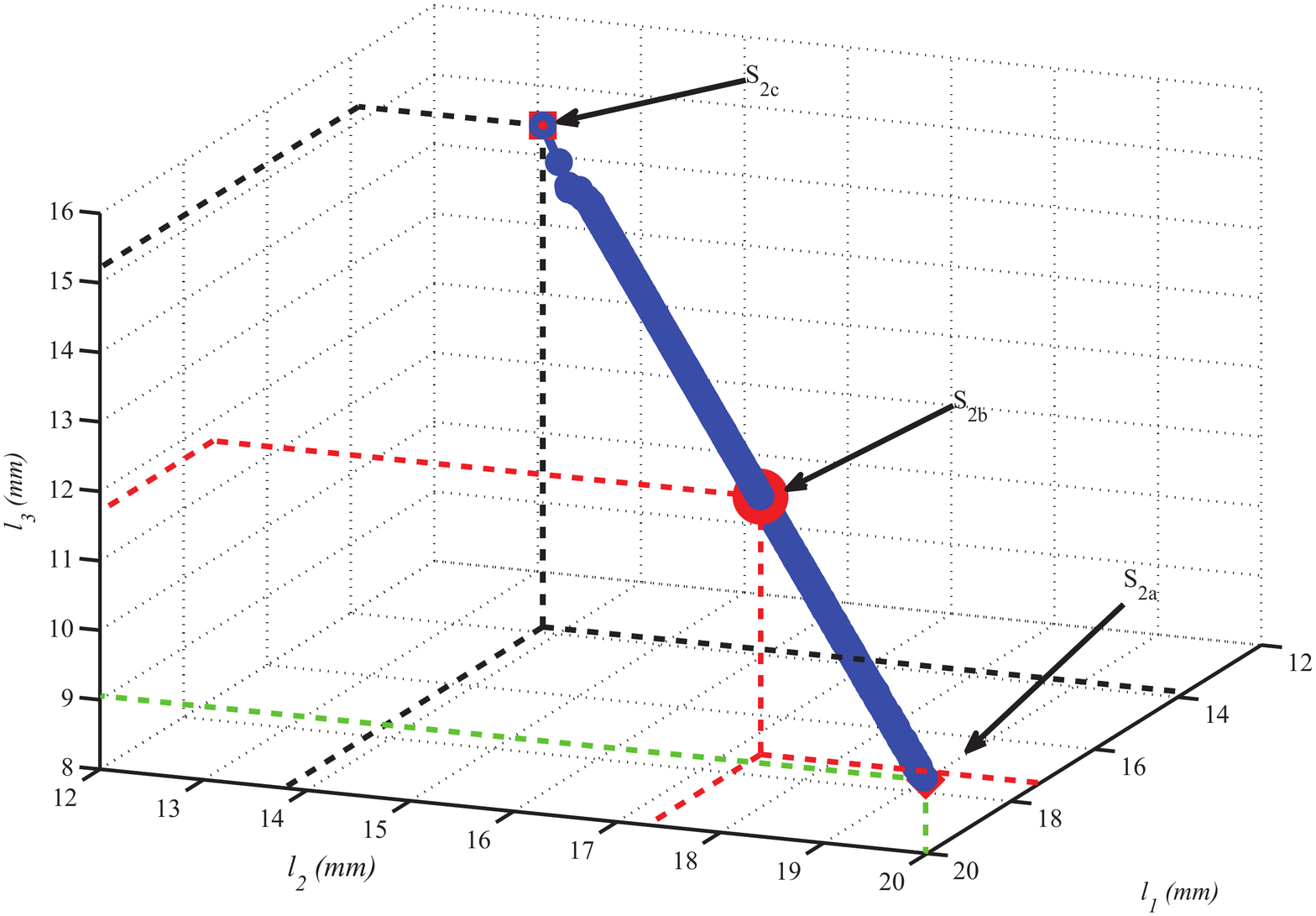}
	\caption{Pareto front for the crank and slider mechanism with 4 bars in the design parameter space. $S_{2a}$ is an optimum efficiency solution , $S_{2b}$ is an intermediate solution and $S_{2c}$ is an optimum size solution .}
	\label{fig:Solution_4bars}
\end{figure}

\begin{figure}[htb!]
	\centering
		\includegraphics[scale=0.70]{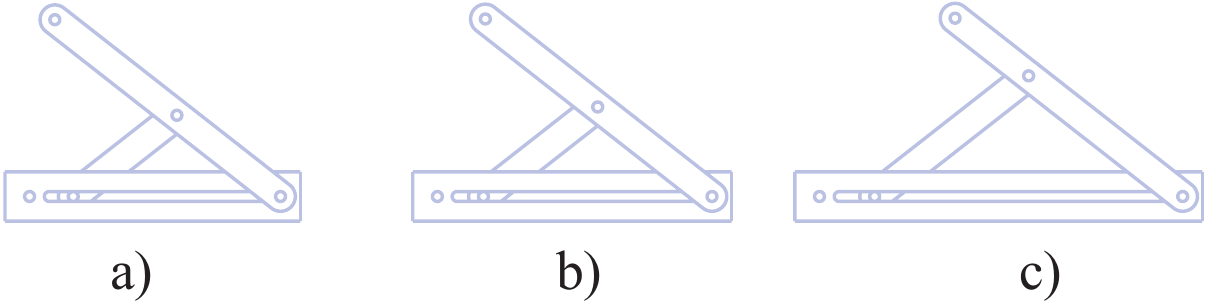}
	\caption{CAD model of the three solutions from the Pareto front for the Crank and slider mechanism with 4 bars a) the optimum size solution, b) the intermediate solution, c) the optimum efficiency solution.}
	\label{fig:CAD_3barres}
\end{figure}

From the Pareto front, we can extract three values:
\begin{itemize}
 \item optimum value of $\Delta x = 25.4 mm$ with $\eta_f =0.52 $: 
\end{itemize}
\begin{equation}
 l_1=13.8 {\rm ~mm}, l_2=13.8 {\rm ~mm}\text{ and } l_3=15.2 {\rm ~mm}\text{ ;}
\end{equation}
\begin{itemize}
 \item optimum value of $\eta_f =0.75 $ with $\Delta x = 35.0 mm$:
\end{itemize}
\begin{equation}
l_1=20.0 {\rm ~mm}, l_2=20.0 {\rm ~mm}\text{ and }l_3=9.0 {\rm ~mm}\text{ ;}
\end{equation}
\begin{itemize}
 \item intermediate solution $\Delta x = 30.3 mm$ and $\eta_f =0.65 $:
\end{itemize}
\begin{equation}
 l_1=17.3 {\rm ~mm}, l_2=17.3 {\rm ~mm}\text{ and }l_3=11.7 {\rm ~mm}\text{ .}
\end{equation}
Figure \ref{fig:Solution_4bars} depicts the evolution of the design variables and Figure~\ref{fig:CAD_3barres} the CAD model associated with the optimal value of $\Delta x$ and $\eta_f$ and an intermediate solution.
We note that $l_1=l_2$ and the Pareto front is defined by:
\begin{equation}
	l_1-2\,l_2-l_3+28.99=0
\end{equation}
The Pareto front is a straight line in the design parameter space of mechanism. The size of the mechanism is proportional in the length of $l_1$ and $l_2$. The transmission force efficiency is proportional in the length of $l_3$ because there is a lever arm between $l_1$ and $l_3$. If $l_1$ is smaller than $l_3$, the size of the mechanism is favoured. If $l_3$ is smaller than $l_1$, the transmission force efficiency is favored.
\subsection*{Crank and slider mechanism with 6 bars}
\begin{figure}[htb!]
	\centering
		\includegraphics[width=8.5cm]{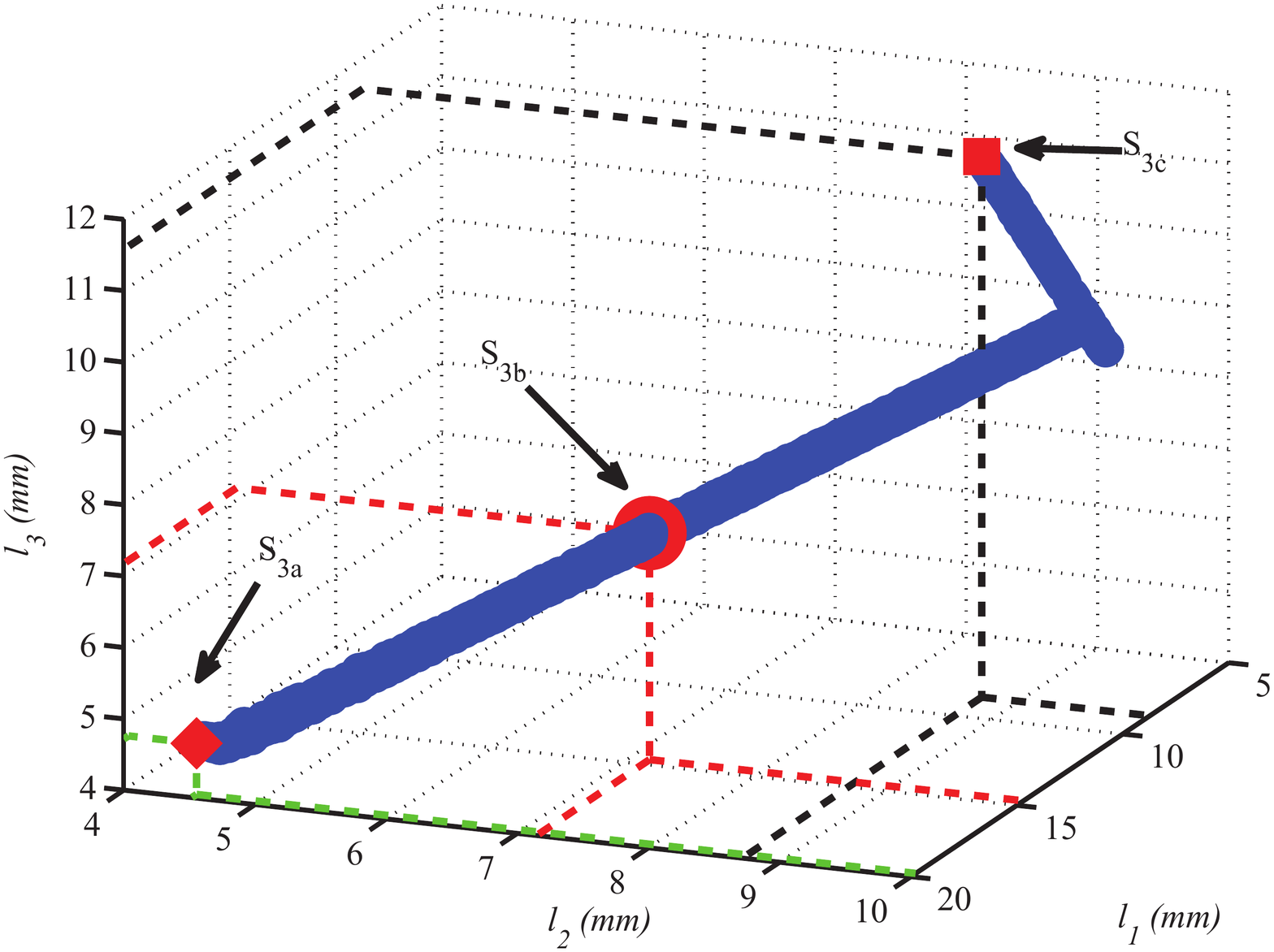}
	\caption{Pareto front for the crank and slider mechanism with 6 bars in the design parameter space. $S_{3a}$ is an optimum efficiency solution , $S_{3b}$ is an intermediate solution and $S_{3c}$ is an optimum size solution .}
	\label{fig:Solution_6bars}
\end{figure}

\begin{figure}[htb!]
	\centering
		\includegraphics[scale=0.70]{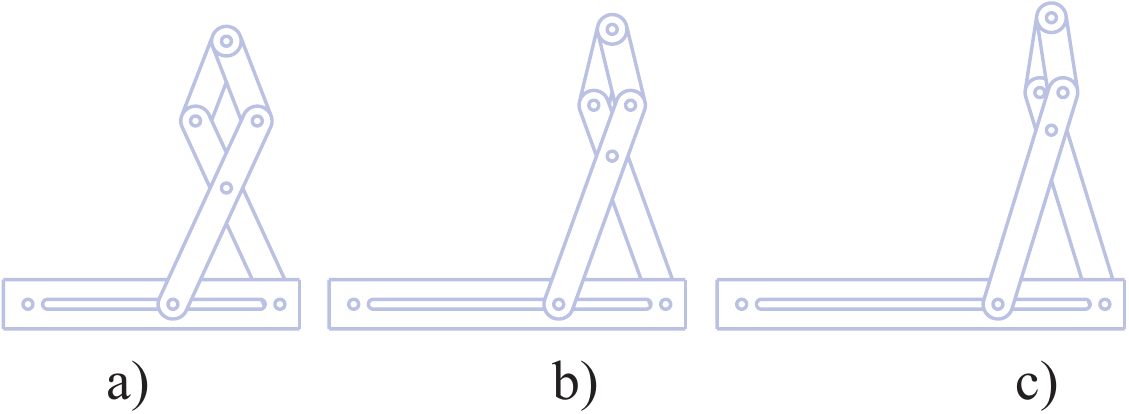}
	\caption{CAD model of the three solutions from the Pareto front for the Crank and slider mechanism with 6 bars a) the optimum size solution, b) the intermediate solution, c) the optimum efficiency solution.}
	\label{fig:CAD_6barres}
\end{figure}

From the Pareto front, we can extract three values:
\begin{itemize}
 \item optimum value of $\Delta x = 16.4 mm$ with $\eta_f =0.30$:
\end{itemize}
\begin{equation}
 l_1=8.8 {\rm ~mm}, l_2=8.8 {\rm ~mm}\text{ and } l_3=11.6 {\rm ~mm}\text{ ;}
\end{equation}
\begin{itemize}
 \item optimum value of $\eta_f =0.75 $ with $\Delta x = 34.9$:
\end{itemize}
\begin{equation}
 l_1=19.8 {\rm ~mm}, l_2=4.6 {\rm ~mm}\text{ and } l_3=4.6 {\rm ~mm}\text{ ;}
\end{equation}
\begin{itemize}
 \item intermediate solution $\Delta x = 25.7 mm$ and $\eta_f =0.56 $:
\end{itemize}
\begin{equation}
 l_1=14.7 {\rm ~mm}, l_2=7.2 {\rm ~mm}\text{ and } l_3=7.2 {\rm ~mm}\text{ .}
\end{equation}
Figure \ref{fig:Solution_6bars} depicts the evolution of the design variables and Figure~\ref{fig:CAD_6barres} the CAD model associated with the optimal value of $\Delta x$ and $\eta_f$ and an intermediate solution.
The Pareto front is defined by:
\begin{equation} 	
-l_1-l_2-l_3+29.04=0 
\end{equation}
The first line of the Pareto front is without constraint $g_9$ and $l_2=l_3$. The second line of the Pareto front with constraint $g_9$ and $l_1=l_2$. The two lines are orthogonal.
If we want to maximize the transmission force efficiency, we are in the case $l_2$ = $l_3$ ($S_{3a}$), so the ratio of the lever arm depends ${l_1}/{l_2}$.
If we want to minimize the size of the mechanism, it is in the case $l_1=l_2$ ($S_{3c}$), so the ratio of the lever arm depends ${l_2}/{l_3}$.
\section*{CONCLUSIONS}
In this article, we have presented a multi-objective optimization problem to design the leg mechanisms of a pipe inspection robot. Three closed loop mechanisms are candidate to be used for the locomotion of this robot.
We have defined two objective functions and a set of constraint equations related to the kinematic behavior.
A genetic algorithm is used to solve this problem with discrete and continuous variables.
The optimization is made for each mechanism as three separate problems to explain the Pareto front obtained by the multi-objective problem resolution. We find out that the {\it slot-follower mechanism} is the best solution for a transmission force efficiency greater than 60\% and the {\it Crank and slider mechanism with 6 bars} is the best for the other cases.
We note that, for our constraint equations, the crank and slider mechanisms have similar performances. 
We select a S1a solution of slot-follower mechanism.
The size of the mechanisms  changes little between different solutions.
The transmission force efficiency significantly changes between 60 \% and 125 \%.
In a practical case, there will be at least 3 legs by groups. But The  contact forces of legs is influenced by  the  cantilever  between points $0$, $P$ and between two groups of legs. 
Similarly, dynamic phenomena are not taken into account in this paper.
The transmission force efficiency is lower than for the slot-follower mechanism. Conversely, the slot-follower mechanism may be more difficult to build because of the passive prismatic and friction can reduce its efficiency. The mechanical strength of axes mechanisms may limit the transmitted power. 
 In future works, the dynamic model with friction parameters will be used in locomotion phases to compare the efficiency of these mechanisms. Several locomotion pattern will be tested in our simulator to ensure contact between the robot and the inside surface of the pipe.
\section*{ACKNOWLEDGMENT}
The work presented in this paper was partially funded by the AREVA company by and a grant of the Ecole des Mines de Nantes, France.
\bibliographystyle{unsrt}
\bibliography{library,library2,library3}
\end{document}